\documentclass[10pt]{article}
\usepackage[top=1  in, bottom=1 in, right=1  in, left=1 in]{geometry}
\usepackage[hidelinks,colorlinks=true,linkcolor=blue,citecolor=blue]{hyperref}

\usepackage{subfigure}
\usepackage{wrapfig,lipsum,booktabs}
\usepackage{microtype}
\usepackage{mathtools}
\usepackage{bbm}
\usepackage{Definitions}
\usepackage{color}
\usepackage{subfigure}
\usepackage{empheq}
\usepackage[ruled,vlined]{algorithm2e}
\usepackage{algorithmic}
\usepackage[utf8]{inputenc} 
\usepackage[T1]{fontenc}    
\usepackage{hyperref}       
\usepackage{url}            
\usepackage{booktabs}       
\usepackage{amsfonts}       
\usepackage{nicefrac}       
\usepackage{microtype}      
\usepackage{cancel}

\begin{document}
\title{Understanding the Spread of COVID-19 Epidemic: A Spatio-Temporal Point Process View}

\author{Shuang Li\thanks{School of Data Science, The Chinese University of Hong Kong, Shenzhen. Email: lishuang@cuhk.edu.cn; chenxinyun@cuhk.edu.cn; fangyixiang@cuhk.edu.cn; songyan@cuhk.edu.cn} \and Lu Wang  \thanks{Microsoft Research, China. Email: joywanglulu@gmail.com. The first two authors contribute equally.} \and Xinyun Chen \footnotemark[1] \and Yixiang Fang \footnotemark[1] 
 \and Yan Song \footnotemark[1] 
 }

\date{}

\maketitle

\begin{abstract}
Since the first coronavirus case was identified in the U.S. on Jan. 21, more than 1 million people in the U.S. have confirmed cases of COVID-19. This infectious respiratory disease has spread rapidly across more than 3000 counties and 50 states in the U.S. and have exhibited evolutionary clustering and complex triggering patterns. It is essential to understand the complex spacetime intertwined propagation of this disease so that accurate prediction or smart external intervention can be carried out. In this paper, we model the propagation of the COVID-19 as spatio-temporal point processes and propose a generative and intensity-free model to track the spread of the disease. We further adopt a generative adversarial imitation learning framework to learn the model parameters. In comparison with the traditional likelihood-based learning methods, this imitation learning framework does not need to prespecify an intensity function, which alleviates the model-misspecification. Moreover, the adversarial learning procedure bypasses the difficult-to-evaluate integral involved in the likelihood evaluation, which makes the model inference more scalable with the data and variables. We showcase the dynamic learning performance on the COVID-19 confirmed cases in the U.S. and evaluate the social distancing policy based on the learned generative model. 
\end{abstract}
\setlength{\abovedisplayskip}{3pt}
\setlength{\abovedisplayshortskip}{3pt}
\setlength{\belowdisplayskip}{2pt}
\setlength{\belowdisplayshortskip}{2pt}
\setlength{\jot}{2pt}
\setlength{\floatsep}{1ex}
\setlength{\textfloatsep}{1ex}
\setlength{\intextsep}{1ex}
\setlength{\topsep}{1ex}
\setlength{\partopsep}{1ex}
\setlength{\parskip}{1ex}

\section{Introduction}
Since the first coronavirus case was confirmed in Washington state on Jan. 21, up to May 21 more than 1.5 million people have confirmed COVID-19 and more than 93,000 people have died from the disease in the U.S.\footnote{Data source: Center for Systems and Engineering at Johns Hopkins University} This infectious respiratory disease has spread rapidly across more than 3000 counties and 50 states, with the exponential growth of confirmed case count in March and with all 50 states reporting cases by March 17. In April, the U.S. became the nation with the most confirmed cases and most deaths globally. On March 15, the Centers for Disease Control and Prevention advised against gatherings of 50 or more people for the next two months, and two of the first U.S. hot spots, Washington state and Illinois, closed all bars and restaurants. On the next day, many cities and states shut down social life and many schools began to close.

The increasing temporal patterns exhibit significant differences county by county, which are influenced by features such as population and location. The three states, New York state, Connecticut, and New Jersey alone, have accounted for about 50\% of all U.S. confirmed cases since March 20. This paper is motivated by modeling and predicting the spread of COVID-19 the exhibits clustering and triggering patterns in time and space. We propose a generative model to track the spread of the disease and directly captures how infections are transmitted. Our model can help understand how one state’s outbreak compares with another’s and provides a simulator to evaluate policy, such as when is the best time to start and ease the restrictions on the social distancing. 

We treat confirmed COVID-19 cases as discrete events, and directly model the transmission of the events by spatio-temporal point processes (STPPs). STPPs model the generative process of discrete events in continuous time and space by intensity function, without the need to divide the space and time into cells ~\cite{moller2003statistical, diggle2006spatio,reinhart2018review}. The occurrence intensity of events is a function of space, time, and history, and explicitly characterizes how the events are allocated over time and space. The propagation of contagious diseases such as COVID-19 often exhibit self-exciting patterns~\cite{reinhart2018review}  that the occurrence of a previous event will boost the occurrence of new events~\cite{mohler2011self,isham1979self,rathbun1996asymptotic} within a region centered around the current location. Existing spatio-temporal self-exciting models require handcrafting the triggering kernel to capture the propagation patterns. The log-Gaussian Cox process, where the log intensity function is a random realization drawn from a Gaussian process~\cite{moller1998log, diggle2013spatial}, 
although flexible, requires a prespecified mean and covariance function to incorporate an accurate prior belief on the spacetime interleaved correlation. Moreover, this model faces challenges to scale with voluminous data like the COVID-19 in the U.S. and is not proper in this setting. 

To alleviate the model-misspecification, we propose a customized imitation learning framework for spatio-temporal point processes. Our policy-like generative models are intensity-free, with the output events (i.e., confirmed case in space and time) directly produced by nonlinear transformations to the history embedding. This generative process mimics the self-exciting mechanism, but the neural-based nonlinear transformations add flexibility to the triggering kernel that can be learned in a data-driven fashion. Furthermore, by incorporating features relative to population, lockdown time, and other spatial and temporal covariates to the intensity function, we add flexibility and interpretability to the model. The learned model can be used to evaluate how population and lockdown time will impact the spread of the virus.

We adopt an imitation learning framework~\cite{abbeel2004apprenticeship,ng2000algorithms,ziebart2008maximum,ho2016generative} to learn the generative model (i.e., policy) by minimizing the discrepancy between the generated events and the observed events, where the learning method is an extension of ~\cite{li2018learning} to the spatio-temporal setting; yet our point process generator is intensity-free. We empirically demonstrate the sound performance of our method in generating and forecasting the confirmed COVID-19 cases in the U.S.

\section{Background: Spatio-Temporal Point Process}
We are interested in learning the generating dynamics of events localized randomly in time and space. Each event is recorded as a tuple $e: = (t, u),$
where $t \in \mathbb{R}^{+}$ is the occurrence time and $u \in \mathbb{S}$~is the occurrence location of the event. A spatio-temporal point processes (STPP) is a random process whose realization consists of an ordered sequence of events, i.e., 
$
\mathcal{H}_t:= \{e_1=(t_1, u_1),  \dots, e_{i}=( t_{i}, u_{i} ) \,|\, t_{i} < t \},
$
where $\mathcal{H}_t$ is the history up to time $t$ and $\mathcal{H}_t$ is $\sigma$-algebra. 

{\bf Conditional Intensity Function.} Denote $N(A)$ as the number of events, such as $e=(t, u)$, falling in the set $A \subset \mathbb{R}^{+} \times \mathbb{S}$. The dynamics of STPP can be characterized by a conditional intensity function, denoted as
\begin{align}
\lambda(t, u \, | \, \mathcal{H}_t) dt du = \mathbb{E}[N(dt \times du ) \,|\, \mathcal{H}_t],
\end{align}
which specifies the mean number of events in a region (i.e., infinitisemal interval and region around $t$ and $u$) conditional on the past. The propagation of contagious diseases often exhibit self-exciting patterns that can be characterized in terms of the conditional intensity function of the form 
\begin{align}\label{eq:exciting}
     \lambda(t, u\, | \, \mathcal{H}_t) = \beta_0(u) + \sum_{i: t_i <t} g(u-u_i, t-t_i)
\end{align}
where $\beta_0(s)$ is the exogenous event intensity that models drive outside the region, and the endogenous event intensity $\sum_{i: t_i <t} g(u-u_i, t-t_i)$ models interactions within the region and $g$ is the triggering kernel. 
Using the chain rule and the conditional density function, we can obtain the joint likelihood for a realization of events $\{e_1=(t_1, u_1), \dots, e_n=(t_n, u_n)| t_n < t\}$ as
\begin{align}\label{eq:likelihood}
\Lcal = \exp\left\{- \int \int_{( 0, t) \times \mathbb{S}} \lambda(\tau, v|\Hcal_{\tau}) d\tau dv \right\} \prod_{i= 1}^n  \lambda(t_i, u_i \Hcal_{t_i} ). 
\end{align}
Suppose a parametric model $\lambda_{\theta}(t, u|\Hcal_t)$ for the conditional intensity has been specified by an unknown parameter $\theta$, then using the maximum-likelihood learning paradigm, one can get an estimation $\hat{\theta}$ by maximizing the likelihood (\ref{eq:likelihood}) in terms of $\theta$. In situations where events exhibit complex or time-evolving patterns, it is difficult to design an expressive intensity function to reflect reality beforehand. Moreover, the evaluation of likelihood (as shown in Eq.~(\ref{eq:likelihood})) involves a three-dimensional integral that usually requires expensive numerical approximation.

\section{Policy-Like STPP Generative Model}\label{sec:policy}
\begin{figure}[htbp]
        \centering
        \includegraphics[width=.50\textwidth]{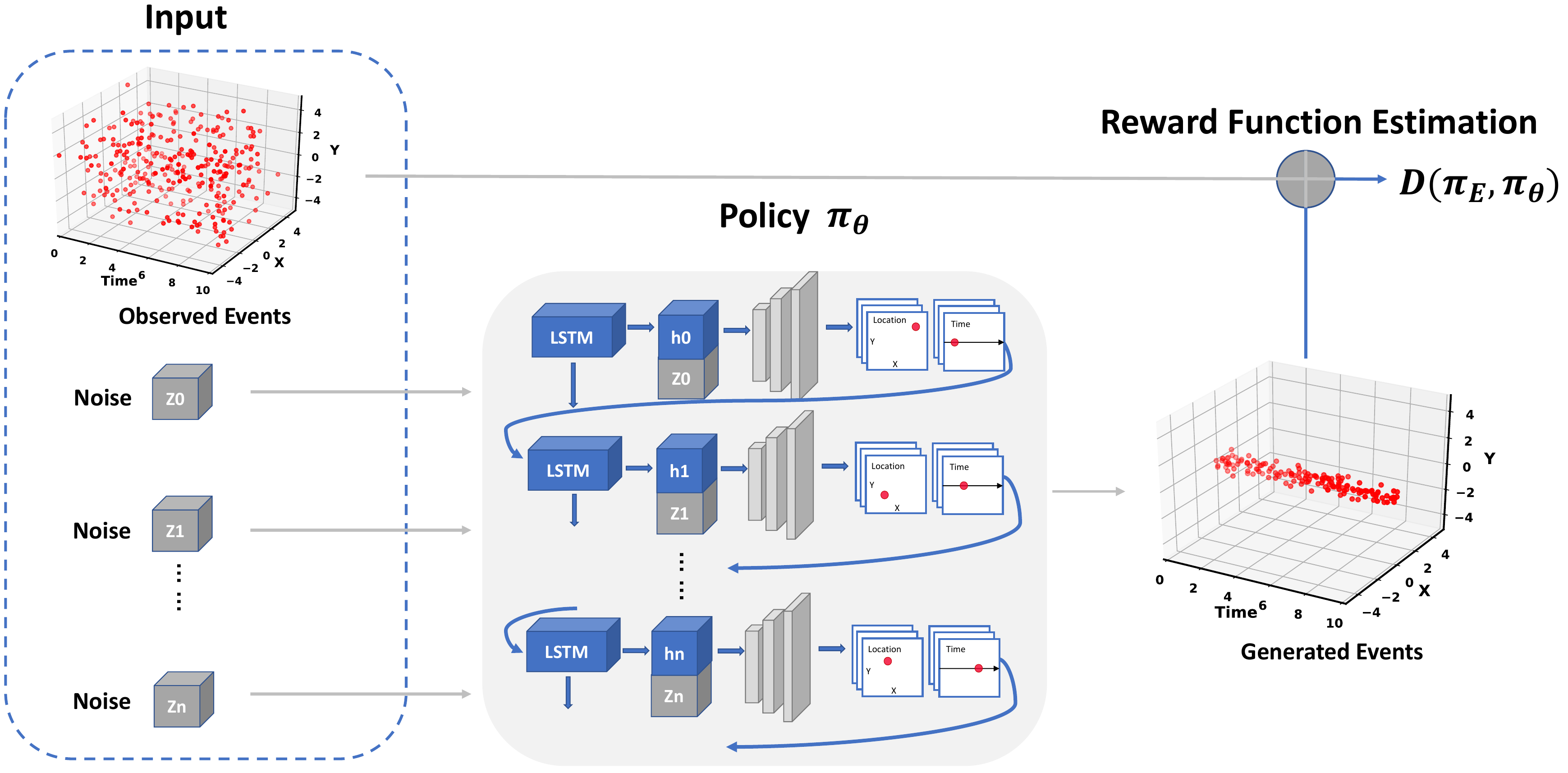}
        \caption{\small{Imitation Learning Framework for Spatio-Temporal Point Processes. }}
        \label{fig:demo}
\end{figure}

We propose a policy-like generative model to mimic the self-exciting triggering patterns of the events as in Eq.~(\ref{eq:exciting}). Our model is {\it intensity-free}, and it is easy to perform model inference, simulate new data, and forecast new events.
Pseudo events (i.e., actions of policy) are sequentially generated from learner policy $\pi_{\theta}(a |\mathcal{S}_t)$, where $\Scal_t$ includes all the previously generated actions. Our generative model is intensity-free in the sense that the output events are directly produced via nonlinear transformations to random noise and history embedding as illustrated in Fig.~\ref{fig:demo}. 
We design policy $\pi_{\theta}(a |\mathcal{S}_t)$ under the principles to: ({\it i}) capture the {\it long-term} and {\it nonlinear} spacetime intertwined dependency of events; ({\it ii}) mimic the {\it doubly stochastic} nature of the triggering spatio-temporal point processes (i.e., the occurrence of an event will influence the occurrence of future events and this is the dynamic of how is the COVID-19 virus spread)~\cite{grandell2006doubly,hawkes1971spectra}; ({\it iii}) and do {\it exploration} over the continuous time and space domains; and ({\it iv}) incorporate features such as population, lockdown time, and other spatial and temporal covariates to add the interpretability to the model. To this end, the policy $\pi_{\theta}(a |\mathcal{S}_t)$  has the following modules:

({\it i}) A recurrent neural network (RNN) unit is to learn an abstract representation from historical events $\mathcal{S}_t$, 
\begin{align}\label{eq:RNNmodel}
h_i  = \psi(Va_i+Wh_{i-1} + B),~~h_0 = 0,
\end{align}
where the input $a = [\tilde{t}, \tilde{u}] $ with $\tilde{t} \in \mathbb{R}^+$ and $\tilde{u} \in \mathbb{R}^2$, the parameter matrices
$V\in \mathbb{R}^{h\times 3}$, $W \in \mathbb{R}^{h\times h}$, $B \in \mathbb{R}^{h}$, the hidden state $h \in \mathbb{R}^{h}$ is an embedding of history, and $\psi$ is an element-wise nonlinear operator. The generated output event $a_t$ will be fed back to the model and serves as the input to trigger (or inhibit) the occurrence of the next event. A random noise vector is injected as part of the inputs to stimulate exploration.

({\it ii}) A multilayer perceptron is applied to noise, hidden state, and static features (i.e., explanatory factors such as population and lockdown time of the city), to generate events, e.g., 
\begin{align}\label{eq:nonlinear}
    a_{i+1} = \sigma( H^{(2)}\sigma(H^{(1)}[h_{i}; z_{i}; f_i] + U^{(1)}) + U^{(2)})
\end{align}
where $h_i$ is the hidden state,  $z_i$ is the noise vector, $f_i$ is the static feature vector, $\sigma(\cdot)$ is an elementwise nonlinear operator, and model parameters $H^{(1)} \in \mathbb{R}^{h'\times (m+h)}$, $U^{(1)} \in \mathbb{R}^{h'}$, $H^{(2)}\in \mathbb{R}^{3\times h'}$, and $U^{(2)}\in \mathbb{R}^3$. The expressive transformation function will represent a rich class of conditional distributions of discrete events~\cite{goodfellow2014generative} that is conditional on history and explanatory variables. Injecting noise to the policy will encourage the {\it exploration} over time and space, and can also be regarded as a ``reparameterization trick" for a random variable, where one substitutes a random variable by a deterministic transformation of a simpler random variable~\cite{kingma2013auto,rezende2014stochastic}. 
\section{Imitation Learning for STPP}
We utilize an imitation learning framework~\cite{li2018learning} to learn our model parameters. In~\cite{li2018learning}, the imitation learning framework is designed for temporal point processes with a prespecified intensity function. We extend this learning framework to the spatio-temporal setting and to the intensity-free models. The imitation learning bypasses the difficult-to-evaluate likelihood and alleviates the model-misspecification at the same time.

In a nutshell, we aim to learn a stochastic policy (learner) $\pi_{\theta}(a |\mathcal{S}_t)$ to mimic the behaviors of the observed events (expert) $\{e_1, e_2, \dots\}$. This is realized by minimizing the discrepancy of the generated events with the observed events evaluated by functions from reproducing kernel Hilbert Space (RKHS). We summarize this imitation learning formulation in Theorem~\ref{theo:formulation} and provide details in Appendix as how to empirically evaluate $D(\pi_E, \pi_{\theta}, \mathcal{F})$ as in Eq.~(\ref{eq:MMD}) by finite samples of event data.
\begin{theorem}\label{theo:formulation}
Let the family of reward function be the unit ball in RKHS $\mathcal{F}$, i.e., $\|r\|_{\mathcal{F}} \leq 1$. Then the optimal policy $\pi_{\theta}(a |\mathcal{S}_t)$ that mimics the observed events (generated by expert $\pi_E$) can obtained by  solving
\begin{align}
\label{eq:MMD0}
\pi^*_{\theta} = \arg \min_{\pi_{\theta} \in \mathcal{G}} D(\pi_E, \pi_{\theta}, \mathcal{F})
\end{align}
where $D(\pi_E, \pi_{\theta}, \mathcal{F})$ is the maximum expected cumulative reward discrepancy between $\pi_E$ and $\pi_{\theta}$, with the expression
\begin{align} \label{eq:MMD}
 D(\pi_E, \pi_{\theta}, \mathcal{F})
 :=\max_{\|r\|_{\mathcal{F}} \leq 1}  \,\, \left( \mathbb{E}_{\xi \sim \pi_E} \left[ \sum\nolimits_{i=1}^{N_T} r( e_i)\right] - \mathbb{E}_{ \eta \sim \pi_{\theta}} \left[\sum\nolimits_{i=1}^{\tilde{N}_T}  r(a_i)\right] \right),
\end{align}
where $N_T$ and $\tilde{N}_T$ are the counts of observed events and generated events within time horizon $T$.
\end{theorem}
This imitation learning formulation accommodates the asynchronous nature of discrete events within a fixed time horizon and a space region. Minimizing $D(\pi_E, \pi_{\theta}, \mathcal{F})$ is equal to minimizing $D^2(\pi_E, \pi_{\theta}, \mathcal{F})$, where the latter is more convenient to use without normalization. Our goal is to learn a policy $\pi_{\theta}$ that the discrepancy is close to zero. Policy parameters $\theta$ will be learned in an end-to-end fashion. 

The gradient of the objective function $D^2(\pi_E, \pi_{\theta}, \mathcal{F}) $ can be backpropagated through the generative policy network. The policy parameters can be optimized by (stochastic) gradient descent method, i.e.,  
\begin{align}
\theta_{k+1} = \theta_k - \alpha_k \nabla_{\theta} D^2(\pi_E, \pi_{\theta}, \mathcal{F})| _{\theta = \theta_k}
\end{align}
where $k$ denotes iteration and $\alpha_k$ is the learning rate. Using the chain rule, 
$
    \frac{\partial D^2(\pi_E, \pi_{\theta}, \mathcal{F})}{ \partial \theta}
    =  \frac{\partial D^2(\pi_E, \pi_{\theta}, \mathcal{F})}{ \partial a} \cdot \frac{\partial a}{ \partial \theta} 
$, where the roll-out samples $a:=\{a_i\}$ are obtained by nonlinear transformations parametrized by $\theta$ and contain the derivative information of $\theta$.
\hspace{-2em}
\section{Generate New Events}
Our generative spatio-temporal point process model is intensity-free. But it is easy to generate new events to {\it predict} ``when" and ``where" the next event will happen, or to {\it evaluate the impact of external variables}. This is realized by generating new action $a$ as the next event from the learned policy
$\pi_{\hat{\theta}}(a |\mathcal{H}_t).$ Specifically, given historical events to predict future events, the sequence of observed {\it real events} up to time $t$,
$\mathcal{H}_t = \{e_1=(t_1, u_1),  \dots, e_{i}=( t_{i}, u_{i} ) \,|\, t_{i} < t \}$ serves as the inputs of the trained recurrent neural network, from which we obtain the {\it last hidden state} $h_i$, i.e., 
\begin{align}\label{eq:generate_event}
& h_i  = \psi(\hat{V} e_i+\hat{W}h_{i-1} +\hat{B}).
\end{align}
Then generate new events via nonlinear transformations to noise, hidden state $h_i$, and features $f_i$, i.e., 
\begin{align}
   z \sim \mathcal{U}(0, 1), \quad \hat{e}_{i+1}= \sigma( \hat{H}^{(2)}\sigma(\hat{H}^{(1)}[h_{i}; z_{i}; f_i] + \hat{U}^{(1)}) + \hat{U}^{(2)}) 
   \label{eq:gennew}
\end{align}
where $\hat{e}_{i+1}$ is the predicted new event. To continue, the new generated event $\hat{e}_{i+1}$ will be fed back the recurrent neural network to update the last hidden state
$
    h_{i+1}  = \psi(\hat{V}\hat{e}_{i+1}+\hat{W}h_{i} + \hat{B})
$
and repeat the previous procedures.

Since we explicitly incorporate external features $f_i$, such as population and lockdown time of a city, to the generative model, the learned model is ready to evaluate how different population or lockdown time of a city will change the propagation of the events. This is realized by generating new events under the new features we want to evaluate. 

\section{Model Validation}
We first empirically validated the dynamic learning and prediction performance of our model on four widely used event datasets. 
We focused on checking whether the learned model can recover the spatial and temporal patterns of the observed events and whether the learned generative model can accurately predict future events. 

{\bf Datasets and Experiment Setup.}
We considered four event datasets: ({\it i}) The UK Car Accidents \footnote{https://data.gov.uk/dataset/cb7ae6f0-4be6-4935-9277-47e5ce24a11f/road-safety-data} that record circumstances of personal injury road accidents in the United Kingdom from the year 2013 to 2015.
({\it ii}) New York City Taxi Trips \footnote{https://www1.nyc.gov/site/tlc/about/tlc-trip-record-data.page} that include NYC Yellow Cab trips in the year 2016. The data were recorded by the NYC Taxi and Limousine Commission (TLC).
({\it iii}) Chicago Crimes\footnote{https://data.cityofchicago.org/Public-Safety/Crimes-2001-to-present/ijzp-q8t2/data} that record the crimes occurred in the City of Chicago from the year 2001 to 2019. The data were collected by the Chicago Police Department's CLEAR (Citizen Law Enforcement Analysis and Reporting) system.
({\it iv}) Worldwide Earthquakes \footnote{https://earthquake.usgs.gov/data/} that record time, location, and magnitudes of all significant earthquakes occurred worldwide from the year 1965-2016. The data were provided by the National Earthquake Information Center (NEIC). 

For these real events, we preserved the time and location (i.e., latitude and longitude) information. We focused on understanding the potentially complex spacetime intertwined dynamics, and their daily or yearly patterns.
In our experiment, the generator is a one layer LSTM with hidden state size 64, and a two-layer MLP with hidden node size 74-32-3. The noise vector is of dimension 10. The generator is optimized by Adam, with a learning rate of $1e$-3. The kernel bandwidth used in evaluating the discrepancy is set to be 1. All the codes and datasets will be released upon the paper is published.
\begin{figure*}[t]
    \centering
    \includegraphics[width=1\textwidth]{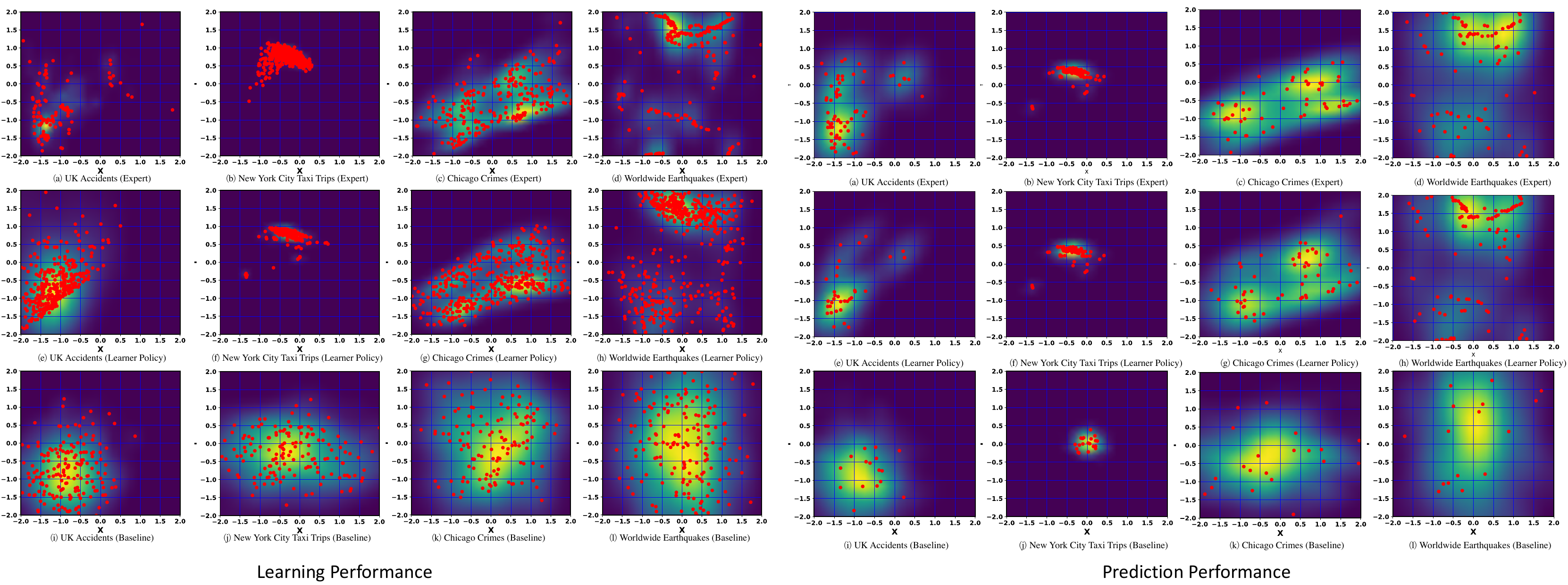}
    \caption{\small{Spatial Distributions of the Observed Events and the Generated Events.}}
    \label{fig:space_learng_prediction}
\end{figure*}
\begin{figure*}[t]
    \centering
    \includegraphics[width=1\textwidth]{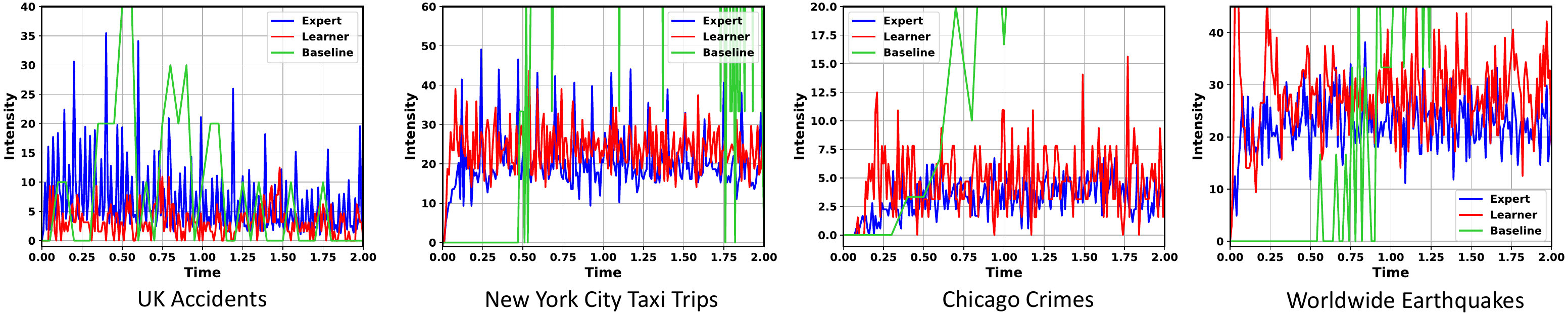}
    \caption{\small{Learning Performance: Intensify Functions of the Observed Events and the Generated Events.}}
    \label{fig:time_learning}
\end{figure*}

\begin{figure*}[t]
    \centering
    \includegraphics[width=0.97\textwidth]{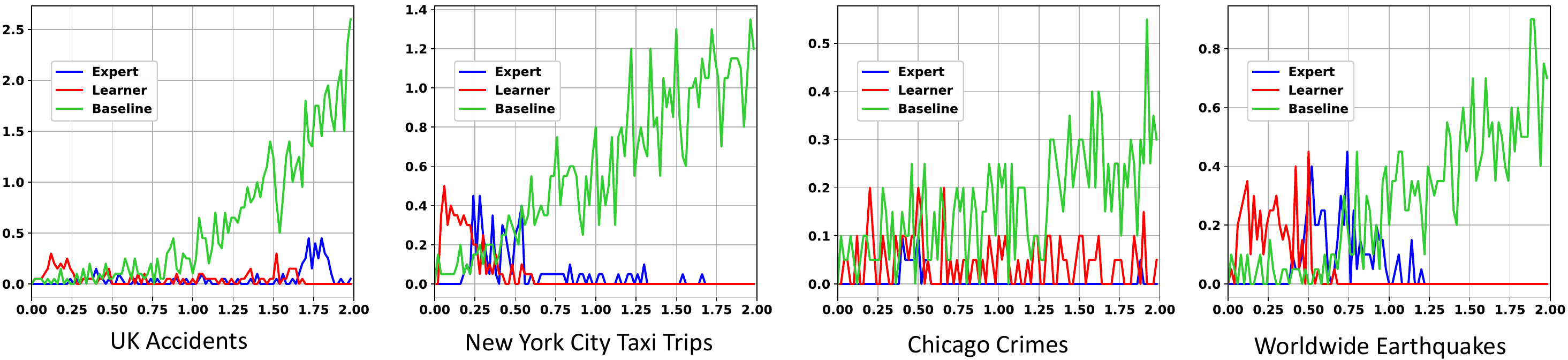}
    \caption{\small{Predication Performance: Intensify Functions of the Observed Events and the Predicted Events.}}
    \label{fig:time_prediction}
\end{figure*}

{\bf Baseline.} The baseline model was designed to share a similar architecture as our generator; but was trained using the MLE. Specifically, the baseline model has the same LSTM unit to embed history as in Eq.~(\ref{eq:RNNmodel}). 
In comparison,
the output events are generated from a prespecified conditional probability density.
We assume  
$t\sim \mbox{Exp}(t| \lambda_{\theta}(h)) \quad \mbox{and} \quad u \sim \mbox{Gaussian}(u | \mu_{\theta}(h), \Sigma_{\theta}(h))$,
which implies the time is generated from an exponential distribution with parameter $\lambda_{\theta}(h) $, and the location is generated from a Gaussian distribution with mean $\mu_{\theta}(h) $ and covariance $\Sigma_{\theta}(h)$. Under these model assumptions, the likelihood of the observed events $\{e_i\}_{i=1, \dots, n}$ can be computed using Eq.~(\ref{eq:likelihood}). Note that the temporal and spatial components of the output events are coupled, due to that the probability distribution parameters all depend on the same hidden state $h$. Parameters of the model were learned via maximizing the likelihood.  

{\bf Learning and Prediction Performance.}
We generated new events from the two well-trained models (i.e., learner policy and baseline), and evaluated the learning performance by checking the recovery results. All pseudo-events were sequentially generated within a predefined time horizon. For the UK Car Accidents, we are interested in the spatio-temporal allocations of the accidents at a daily level, and we chunk this dataset according to days to construct the sequences of events. For the New York City taxi Trips, we also split this dataset according to days and further gleaned the trips that happened between 3:00 pm to 4:00 pm each day. For Chicago Crimes, we split this dataset based on the date to construct the sequences of events, and further focused on the crimes that happened between 9:00 pm to 12:00 pm. For Worldwide Earthquakes where the events are sparse, we split this dataset according to years (since earthquakes are rare events) to construct the sequences of the events, and we aim to model the spatio-temporal patterns of the earthquakes at an annual level.

We linearly scaled the event times and locations (i.e., the scaled times fall in the region of $[0, 2]$ and the scaled latitudes and longitudes fall in the region of $[-2, 2]$ ) so that they displayed in a reasonable range for training stability and visualization purposes. We demonstrated the spatial distributions and the temporal intensity functions of the real events (expert) and the generated events (learner) in Fig.~\ref{fig:space_learng_prediction} (left) and Fig.~\ref{fig:time_learning}. In the figures, we only visualized a batch (i.e., 32) of generated sequences. As demonstrated, our learner policy, which jointly generated the spatial and the temporal components of the events, successfully recovered the underlying patterns of the real events, and especially was able to capture the {\it sparse and irregular patterns} over the space. 

The baseline model, however, failed to capture the fine-grained patterns, which also indicates the conditional density assumption for the emission probabilities is quite restricted. If we used the baseline model to only learn and generate the temporal component of the events (i.e., we ignored the spatial information of the events), the baseline model can accurately recover the temporal patterns(we implemented the experiments but didn't display the results here). But if we let the baseline model jointly learn and generate the spatial and temporal components of the events, then the learning performance significantly degraded -- the difficulties in capturing the spatial patterns will destroy the generative performance for the temporal patterns. To remedy this problem, one may resort to a nonparametric Gaussian mixture model to capture the clustering patterns for the spatial component, but still, this design is not generic and can only work for some specific patterns. Moreover, sophisticated models will lead to a hard to evaluate likelihood. The comparison results showcase the power of our method in learning the complex spatial-temporal patterns of real events

We also evaluated the prediction accuracy of the two-well trained models, i.e., to predict when and where the next event will happen given the observed history. Our model and the baseline models were evaluated by conditioning on the same historical real events, i.e., the same sequence of real observed events was injected to the LSTM as inputs to obtain the last hidden state, and this last hidden state was used as an initial state to generate new events as predictions. The predicted results were demonstrated in Fig.~\ref{fig:space_learng_prediction} (right) and Fig.~\ref{fig:time_prediction} (the time horizon has been scaled to $[0, 2]$), where we visualized a batch of next ten predicted events. The results also demonstrated the superior prediction performance of our model, especially in predicting the event locations in comparison with the baseline.

\section{Case Study: COVID-19}

Next, we demonstrated how to utilize our spatial-temporal point process model to understand the spread of the COVID-19 virus in the U.S., and we especially interested in predicting how the social distancing policy (i.e., city lockdown time) will influence the propagation of the disease.  
\begin{figure}[h!]
    \centering
    \includegraphics[width=0.5\textwidth]{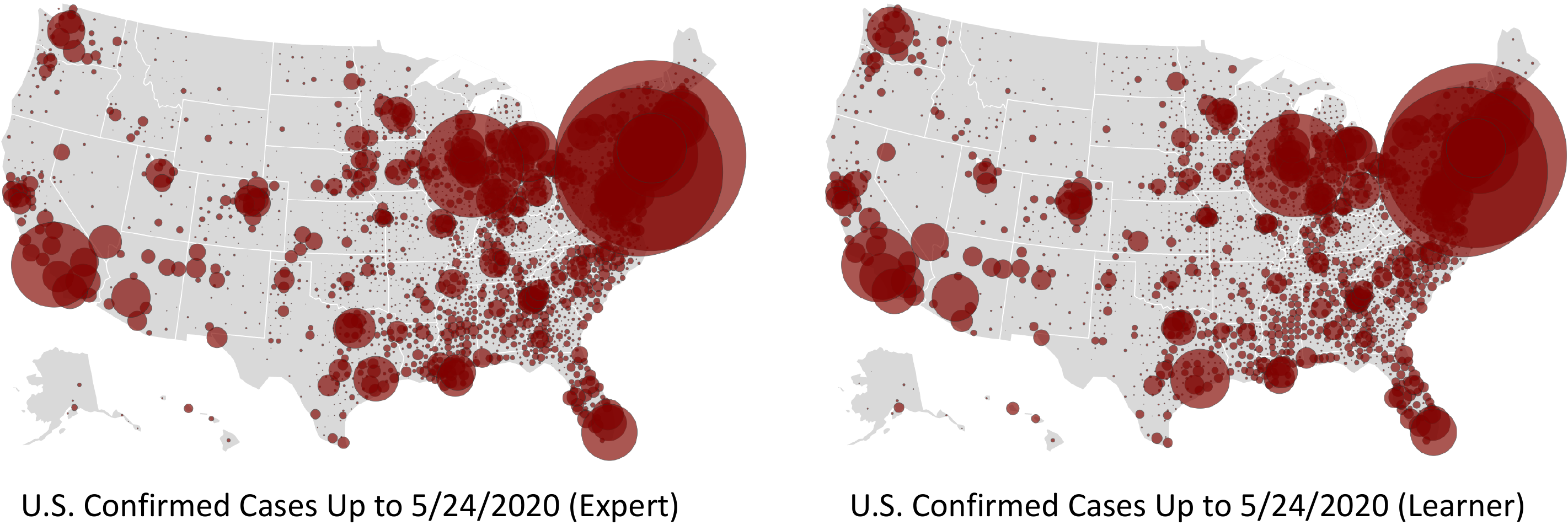}
    \caption{\small{Comparison of the Real and the Generated Cumulative Count of Confirmed COVID-19 Cases.}}
    \label{fig:map}
\end{figure}

{\bf Datasets and Experiment Setup.}
We considered the dataset from the Center for Systems and Engineering at Johns Hopkins University \footnote{https://coronavirus.jhu.edu/}. This dataset daily updates the number of confirmed COVID-19 cases globally. We only focused on the U.S. confirmed cases reported after Jan 21, 2020, till May 24, 2020.  
In this dataset, each county has its report. Yet, the fine-grained location of each confirmed case is not provided. We can only get access to the location of the county. We, therefore, adjust our generative model to account for this dataset -- we use the location information of the county as markers (i.e., static features) of events and only generate the occurrence time of the new confirmed case. The external features, population, lockdown time, together with the county location are fused to our generator to capture how these factors will influence the event triggering dynamics. Unlike the crimes or accident events that usually exhibit repeated daily patterns, the outbreak of diseases has its unique evolutionary progressions. The trajectories of events are nonstationary over time and exhibit different patterns county by county. We treat each county's confirmed cases as a sequence and predict when is the next confirmed case for this county. 

The setup of the models is the same as before. In this context, we carefully construct the dataset as follows. We divide the 3261 counties' sequences into five groups via the sequence length: length $\leq 100$ (\# 2245); length $> 100$ and $\leq 1000$ (\# 796); length $> 1000$ and $\leq 5000$ (\# 155); length $> 5000$ and $\leq 10000$ (\# 39); length $> 10000$ and $\leq 20000$ (\# 18); and 8 counties is with length $>20000$.
We train a model for each group. This is because the counties within the same group share similarities in their virus spread patterns. Short sequences don't have sufficient data to train an individual model. We are pooling the information of other similar counties in the group in the prediction. 
For hot spots, like New York City, the cumulative confirmed count is close to 200,000, we train an individual model. Rather than generating the occurrence time of the next event, we generate the occurrence time when the new cumulative confirmed cases exceed one hundred. 

\begin{figure*}[h!]
    \centering
    \includegraphics[width=1\textwidth]{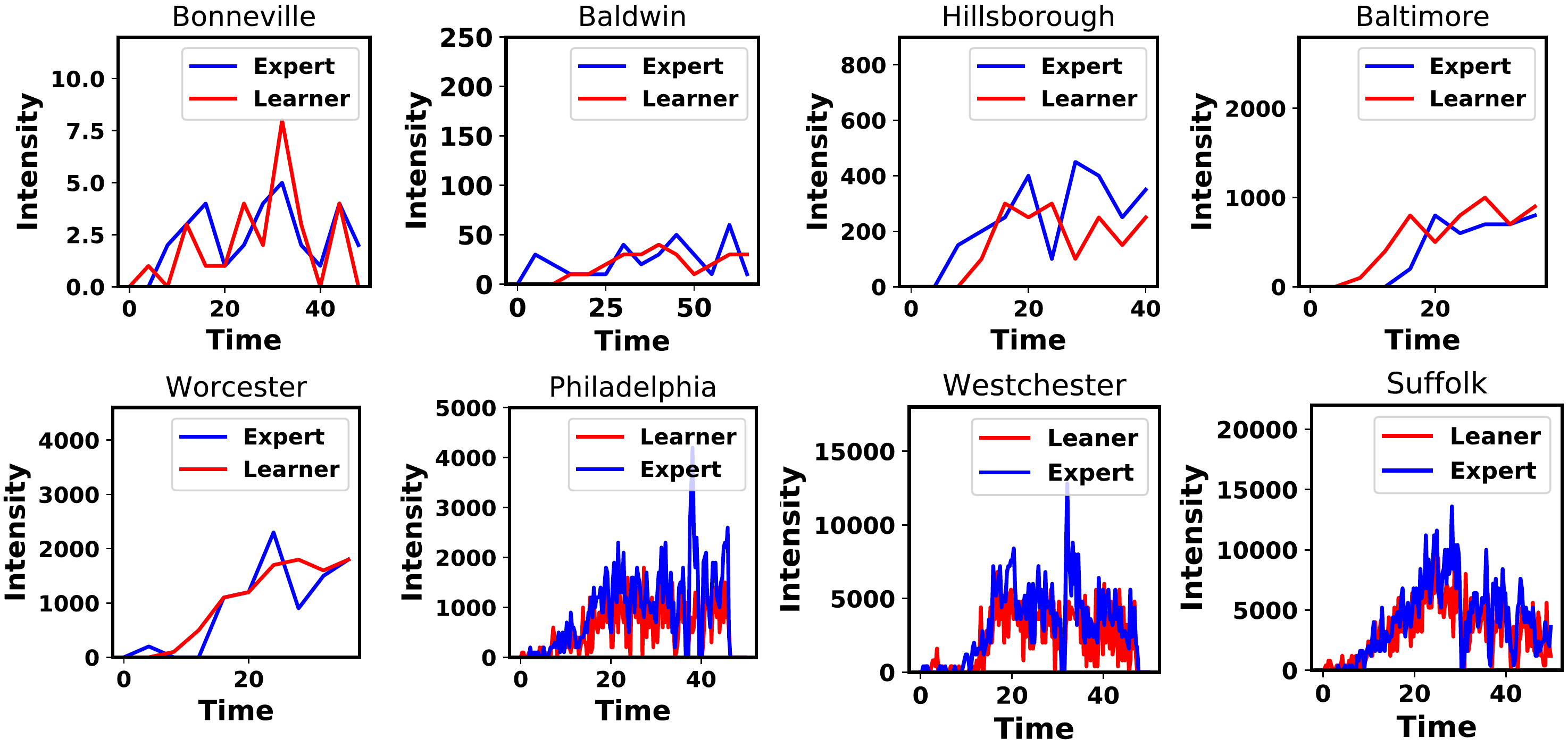}
    \caption{\small{Comparison of the Intensity Functions of The Real and the Generated Confirmed COVID-19 Cases. The Unit of The Time is One Day.}}
    \label{fig:covid_intensity}
\end{figure*}

{\bf Learning and Prediction Performance.} The learning performance of our model on all counties in the U.S. is demonstrated in Fig.~\ref{fig:map}. We generated new events from the learner policy and evaluated the learning performance by checking the recovery results. The results are an aggregation the events in terms of the empirical intensity. The results demonstrate a sound recovery performance of our models. 
We randomly selected eight counties' generating results and demonstrated them in Figure~\ref{fig:covid_intensity} (the measurement unit of the time as shown in the X-axis is one day). We observe that our models can capture the distinct patterns of each county, and the patterns show significant differences across counties. For hot spots, for example, the Suffolk, we can see that the emerging cases grow quickly, with a curve trending sharply upward. As new cases slow, the curve bends toward horizontal, showing that the state’s outbreak may be leveling off. However, this does not mean the number of cases has stopped growing but indicates that the rate of growth has slowed, which could signify that social distancing measures are having an effect. The results show that our model can accurately recover the spread of events with various lengths and trends. Our model especially demonstrates better performance on hot spots (i.e., Philadelphia, Westchester, and Suffolk) mainly due to the sufficient training data. For short sequences, although we adopt pooling techniques, it is still challenging to capture the distinct patterns of each county. 
\begin{table*}
\centering
\caption{Mean value of the predicted total confirmed case on 5/24/20 with early and late lockdown time on four counties. }
\begin{tabular}{cccc} \toprule
County Name&Real Lockdown&Early Lockdown&Late Lockdown\\ \midrule
New York&342240&$292820\pm5641$&$348700\pm6225$\\ 
Los Angeles&39960& $39420\pm1180$&$41230\pm1210$\\ 
Cook &31700&$25600\pm410$&$42365\pm965$\\ 
Middlesex&10431&$10378\pm264$&$12739\pm358$\\ 
\midrule
\end{tabular}
\label{tab:predict}
\end{table*}

We are interested in evaluating how the lockdown time of each county will influence the spread of the diseases. Given a learned model, we generated events by tuning the lockdown time. We predict the trajectories of the events given a different lockdown time: one week earlier and one week later than the real lockdown time.  
The predicted intensity functions are shown in Figure 7 and Tabel~\ref{tab:predict}. We observe that, for all these four counties, the number of events with delayed lockdown time is slightly larger than the counterparts with real lockdown time; the number of events with earlier lockdown time is slightly smaller than the counterparts with real lockdown time. With an earlier lockdown time, Los Angeles and Middlesex show similar trends compared to the real cases. The reason would be that the lockdown time does not make a significant effect on these counties in the early stage of the epidemic. For New York and Cook, however, the early lockdown time shows a significant reduction in the number of total confirmed cases. The results suggest that New York and Cook could control the disease better if they take an earlier lockdown time.

\begin{figure*}[h!]
    \centering
    \includegraphics[width=1\textwidth]{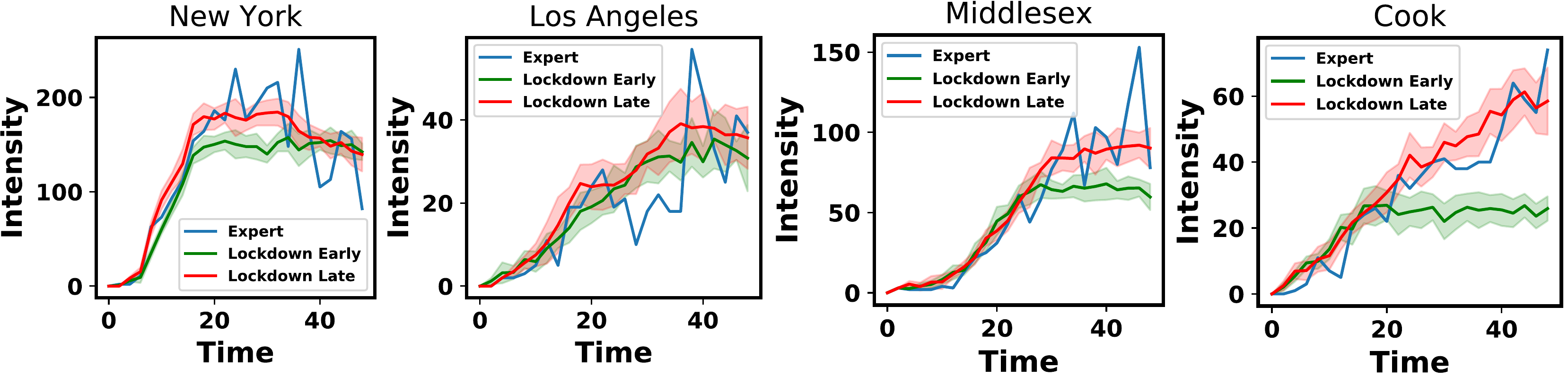}
    \caption{\small{The Predicted Propagation of Events If The Lockdown Time is One Week Earlier or One Week Later Than The Real Lockdown Time. The solid line in red and green indicates the mean of 10 predicted sequences and shade fields indicates the standard deviation. }}
    \label{fig:lockdown}
\end{figure*}
\section{Discussion}
In this paper, we proposed an intensity-free spatio-temporal point processes model and train the model using an imitation learning framework. This learning method bypasses the evaluation of the likelihood function, which has the potential to achieve a good balance between model flexibility and computational burden. We empirically showed the superiority of the proposed method in recovering the complex dynamics of real events and forecasting new events. We especially use COVID-19 as a case study and utilize our model to understand the spread of this virus.

\bibliography{sample-sigconf}
\bibliographystyle{plain}
\newpage
\appendix
\section{Supplementary Materials}

In the following, we summarize the imitation learning formulation in Theorem~\ref{theo:formulation} and provide details as how to empirically evaluate $D(\pi_E, \pi_{\theta}, \mathcal{F})$ as in Eq.~(\ref{eq:MMD}) by finite samples of event data.

\subsection{Imitation Learning Formulation}
In a nutshell, we aim to learn a stochastic policy defined as
$$\pi_{\theta}(a |\mathcal{S}_t)$$
to mimic the behaviors of the observed events. For the learner policy $\pi_{\theta}(a |\mathcal{S}_t)$, we let the state
\begin{align*}
\mathcal{S}_t := \{a_1=(\tilde{t}_1, \tilde{u}_1), \dots, a_{i}=( \tilde{t}_{i}, \tilde{u}_{i} ) \,|\, \tilde{t}_{i} < t \}
\end{align*}
refer to the generated pseudo events up to time $t$, and let action $a_i=(\tilde{t}_i, \tilde{u}_i)$ indicate the generated pseudo event. The next pseudo event (i.e., new action $a_{i+1}$) is taken by policy $\pi_{\theta}(a |\mathcal{S}_t)$ conditional on current state. Then the new state is updated as $$S_{t_{i+1}}=\{a_1, \dots, a_i, a_{i+1}\}.$$
Under the imitation learning framework, we define $\pi_E$ as the expert's policy, which characterizes the dynamics of observed events and we assume the observed events are generated from expert policy $\pi_E$. For spatio-temporal point processes, one can think of $\pi_E:=p( e_i \,|\, \mathcal{H}_{t_i})$ is the conditional density for the next event. We aim to learn a learner policy $\pi_{\theta}:=\pi_{\theta}(a |\mathcal{S}_t)$ to imitate $\pi_E$. 


The imitation learning requires first to learn an optimal reward function $r^*$ from data, via solving
\begin{align}\label{eq:IRL}
D(\pi_E, \pi_{\theta}):=\,\,\max_{r\in \mathcal{R}} \, \, \left(   \mathbb{E}_{\pi_E} \left[\sum_{i=1}^{N_T}r(e_i)\right] - \max_{\pi_{\theta} \in \mathcal{G} } \mathbb{E}_{\pi_{\theta}}\left[ \sum_{i=1}^{\tilde{N}_T}r(a_i)\right]  \right),
\end{align}
where $\mathcal{G}$ is the family of all candidate learner policies $\pi_{\theta}$, ${N}_T$ is the total number of observed events up to time $T$, and $\tilde{N}_T$ is the total number of generated pseudo events up to time $T$.

Given unknown reward function and only expert's sequences of events, the reward function is learned (as in Eq.~(\ref{eq:IRL})) under the principle that expert policy should be uniquely optimal given the reward. In other words, the expert performs better than any other policies $\pi_{\theta }\in \mathcal{G}$ given the reward. This reward learning procedure is also called inverse reinforcement learning. However, the above inverse reinforcement learning formulation is challenging to solve, which requires to solve a reinforcement learning problem in an inner loop and is time-consuming and resource intensive. We will discuss how to simply this formulation and directly obtain an analytical optimal reward function later. 

Given the learned optimal reward $r^*$, the optimal learner policy is obtained by maximizing the cumulative reward of the actions given a finite time horizon $T$
\begin{align} \label{eq:RL}
\vspace{-3mm}
\pi_{\theta}^* = \arg \, \max_{\pi_{\theta} \in \mathcal{G}}~~ J(\pi_{\theta}):= \mathbb{E}_{\pi_{\theta}}\left[\sum_{i=1}^{\tilde{N}_T}r^*( a_i)\right].
\end{align}
In summary, the overall generative adversarial imitation learning framework is illustrated in Fig.~\ref{fig:demo}. Pseudo-events are sequentially generated from policy $\pi_{\theta}$. The discrepancy $D(\pi_E, \pi_{\theta})$ between the generated events with the observed events is evaluated by a reward function. The policy and the reward are jointly learned from data, and the reward function will iteratively guide the policy to improve the sample quality until the sampled events and the real events are indistinguishable.

\subsection{Reward}\label{sec:reward}
Under the imitation learning framework, the reward function and the policy are jointly learned from the data as in Eq.~(\ref{eq:IRL}). The reward function is learned under the principle that expert policy should be uniquely optimal. In this two-player game as shown in Eq.~(\ref{eq:IRL}), the reward that quantifies the discrepancy between $\pi_E$ and $\pi_{\theta}$ is updated by considering the worst-case, and the optimal policy $\pi_{\theta}^*$ aims to close this gap.

The function class for reward should be carefully chosen. On the one hand, we want the reward function class to be sufficiently expressive so that it can represent the reward function of various shapes. On the other hand, it should be restrictive enough to be efficiently learned with finite samples. With the above competing considerations, we choose the reward function class to be the unit ball in RKHS $\mathcal{F}$, i.e., $\|r\|_{\mathcal{F}} \leq 1$. An immediate benefit of this function class is that we can show the optimal policy can be directly learned via a minimization formulation (provided in Theorem~\ref{theo:formulation}) instead of the original minimax formulation as in Eq.~(\ref{eq:IRL}). 

A sketch of proof is provided as follows. We will start from Eq.~(\ref{eq:IRL}) to derive the results in Theorem~\ref{theo:formulation}. Fixing time horizon $T$, let $\xi$ stand for the sequence of observed events up to $T$, i.e., 
$ \xi = \{ e_1, e_2, \dots, e_{N_T}\}$, and we use $\eta$ to stand for the sequence of generated events, i.e.,  
$\eta = \{ a_1, a_2, \dots, a_{\tilde{N}_T}\}$.
For short notation, we denote
\begin{align}
\underbrace{ \phi(\xi) := \int \int_{[0, T)\times \mathbb{S}}  k(e, \cdot) dN_e}_{\text{feature mapping from data space to R} }
\end{align}
\begin{align}
\text{and} \underbrace{ {\mu_{\pi_{E}}} :=  \mathbb{E}_{\xi \sim \pi_{E}}\left [  \phi(\xi)  \right]  }_{\text{mean embeddings of the intensity function in RKHS}} 
\end{align}
where $dN_e$ denotes the counting process associated with sample path $\xi$, and $k(e, e')$ is a universal RKHS kernel. Similarly, we can define $\phi(\eta)$ and $\mu_{\pi_{\theta}} $.
Then using the reproducing property, we write the cumulative reward for the learner policy as
  \begin{align*}
 J(\pi_{\theta}):
 & = \mathbb{E}_{\eta \sim \pi_{\theta}}\left[\int \int_{[0, T)\times \mathbb{S}}  \langle r, k\left(a, \cdot \right) \rangle_{\mathcal{F}} dN_{a}\right]=  \langle r,  \mu_{\pi_{\theta}} \rangle_{\mathcal{F}}.
\end{align*}
Similarly, we obtain $J(\pi_E)=\langle r, \mu_{\pi_E} \rangle_{\mathcal{H}}$. From (\ref{eq:IRL}), $r^*$ is solved by
       \begin{align*}
   \max_{\|r \|_{\mathcal{F}}\leq 1} \, \,  \min_{\pi_{\theta} \in \mathcal{G}} \,  \langle r,  \mu_{\pi_{E}} -  \mu_{\pi_{\theta}}  \rangle_{\mathcal{F}} 
      =& \min_{\pi_{\theta} \in \mathcal{G}} \, \,  \max_{\|r \|_{\mathcal{F}}\leq 1} \langle r,  \mu_{\pi_{E}} -  \mu_{\pi_{\theta}}  \rangle_{\mathcal{F}}  \\
        =  & \min_{\pi_{\theta} \in \mathcal{G}} \| \mu_{\pi_{E}} -  \mu_{\pi_{\theta}}  \|_{\mathcal{F}} ,
       \end{align*}
       where the first equality is guaranteed by the {\it minimax theorem}, and 
\begin{align}\label{optreward}
r^*(\cdot|\pi_E, \pi_{\theta})=\frac{\mu_{\pi_E}-\mu_{\pi_{\theta}}}{\| \mu_{\pi_E}-\mu_{\pi_{\theta}}\|_{\mathcal{F}}} \propto  \mu_{\pi_E}-\mu_{\pi_{\theta}} .
\end{align}
In this way, we convert the original mini-max formulation for solving $\pi^*_{\theta}$ to a simple {\bf minimization} problem, which will be more efficient and stable to solve in practice. We summarize the formulation in Theorem~\ref{theo:formulation}.

\subsection{Finite Sample Estimation} 
\begin{figure}[h!]
        \centering
        \includegraphics[width=.45\textwidth]{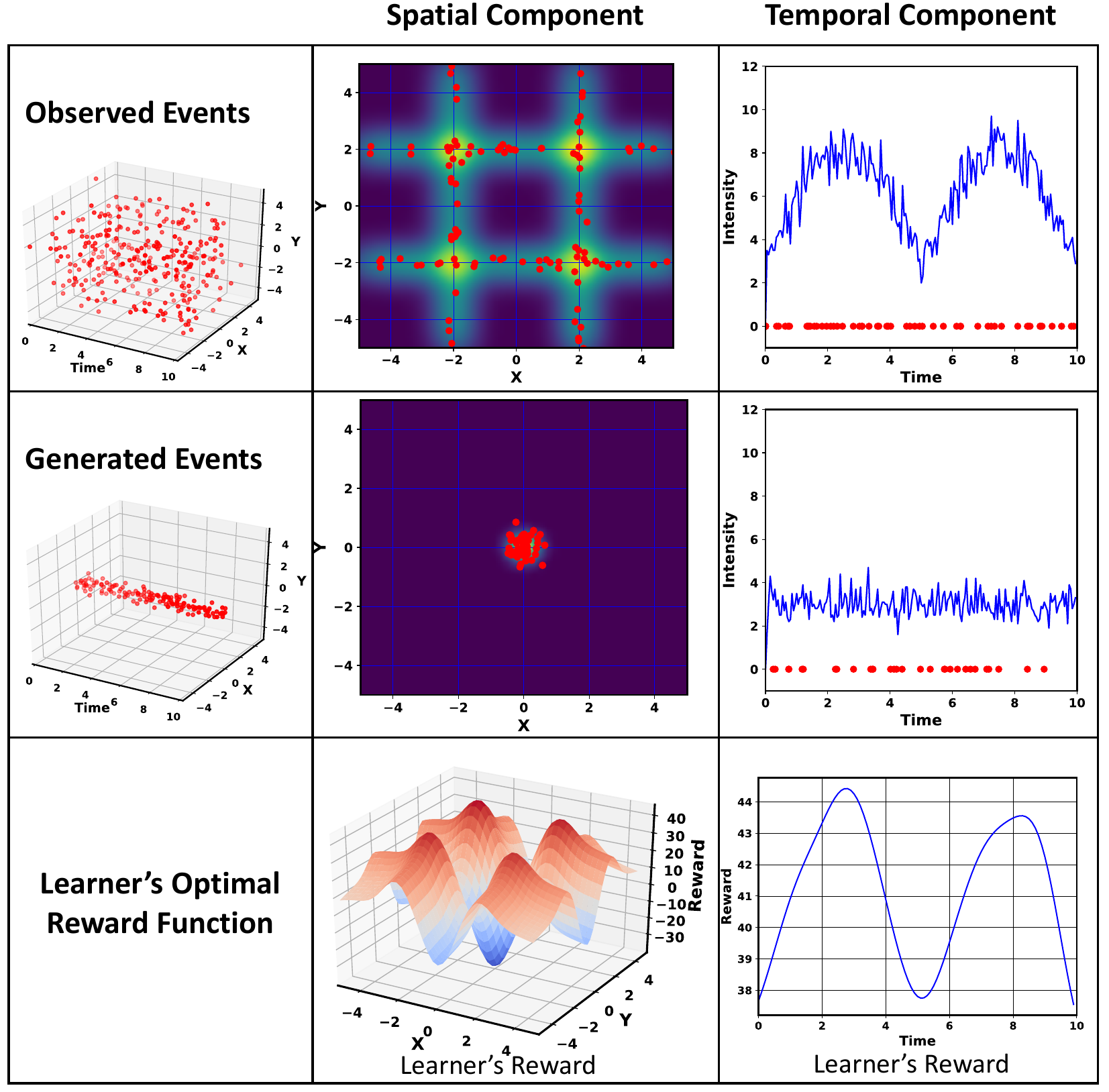}
        \caption{Illustration of Reward Function. }
        \label{fig:demo_reward}
\end{figure}

The optimal reward has an analytical expression as in Eq.~(\ref{optreward}), which can be estimated from finite samples. Given $L$ trajectories of expert point processes, and $M$ trajectories of events generated by $\pi_{\theta}$, the mean embedding $\mu_{\pi_E}$ and $\mu_{\pi_{\theta}}$ can be estimated by their respective empirical mean:
\begin{align}\label{eq:mean_embedding_est}
 \hat{\mu}_{\pi_E} =  \frac{1}{L}\sum_{l=1}^L \sum_{i = 1}^{N^{(l)}_T} k(e^{(l)}_i, \cdot)\quad \text{and} \quad 
 \hat{\mu}_{\pi_{\theta}} =   \frac{1}{M} \sum_{m=1}^{M}   \sum_{i = 1}^{N^{(m)}_T} k(a^{(m)}_i, \cdot).
 \end{align}
Then for any $a:=(t, u)$ where $t \in [0,T)$ and $u \in \mathbb{S}$, the estimated optimal reward evaluated at point $a$ is (without normalization) is 
 \begin{align}\label{eq:rewardest}
 \hat{r}^*(a) \propto & \frac{1}{L}\sum\nolimits_{l=1}^L \sum\nolimits_{i = 1}^{N^{(l)}_T} k(e^{(l)}_i, a ) 
  -  \frac{1}{M} \sum\nolimits_{m=1}^{M}   \sum\nolimits_{i = 1}^{N^{(m)}_T} k(a^{(m)}_i, a) .
\end{align}
The optimal reward is a three-dimensional function over time and space, which recognizes the differences between the generated events and the observed events. This can be further illustrated in Fig.~\ref{fig:demo_reward}. Given the observed events and the current generated events (we visualize the distribution of the events over the spatial component and the occurrence intensity over the temporal component respectively), we plot the estimated optimal reward function using Eq.~(\ref{eq:rewardest}). Since the reward is a three-dimensional function, in the figure, we fix $t=t_0$ (where $t_0=5$) and plot $ \hat{r}^*(x, y, t_0)$ over the $XY$ space; and we fix $u = (x_0, y_0)$ (where $(x_0, y_0)=(0,0)$) and demonstrate $\hat{r}^*(x_0, y_0, t)$. As illustrated, the computed reward function indicates the differences between the observed events' occurrence intensity (expert) and the generated events' occurrence intensity (learner) -- the reward has a high value where the expert's intensity is greater than the learner's intensity, and has a low value where the expert's intensity is smaller than the learner's intensity. In this way, the reward function will guide the learner to mimic the expert in training via obtaining a policy to maximize the cumulative reward.

\subsection{Kernel Choice} 
The unit ball in RKHS is dense and expressive. Fundamentally, our proposed framework and theoretical results are general and can be directly applied to other types of kernels. For example, we can use the Mat\'{e}rn kernel, which generates spaces of differentiable functions known as the Sobolev spaces. In later experiments, we have used Gaussian kernel and obtained promising results. As for future work, the kernel function can also be learned to maximize the distinguish power.

\subsection{Learning Algorithm}
Minimizing $D(\pi_E, \pi_{\theta}, \mathcal{F})$ is equal to minimizing $D^2(\pi_E, \pi_{\theta}, \mathcal{F})$, where the latter formulation is more convenient to use without the need to perform normalization. Our goal is to learn a policy $\pi_{\theta}$ that the discrepancy is close to zero, i.e., 
\begin{align}\label{eq:obj_sq}
\min_{\pi_{\theta}} \quad D^2(\pi_E, \pi_{\theta}, \mathcal{F})
\end{align}
with the corresponding optimal reward function has an analytical and nonparametric expression $r^*= \mu_{\pi_E}-\mu_{\pi_{\theta}}$.

In the original imitation learning formulation as shown in Eq.~(\ref{eq:IRL}), which is a mini-max game solved by alternating the minimization subproblem for the policy and the maximization subproblem for the reward. The optimization procedure should be carefully scheduled to escape bad local optima, and this is in particular for a non-convex game. In our framework, we propose a neural-based generative policy for spatio-temporal point process to gain model expressiveness; and we introduce a nonparametric reward to make our model parsimonious. Overall, our framework is lightweight.

Policy parameters will be learned via solving (\ref{eq:obj_sq}) in an end-to-end fashion. The gradient of the objective function $D^2(\pi_E, \pi_{\theta}, \mathcal{F}) $ can be backpropagated through the generative policy network. The policy parameters can be optimized by (stochastic) gradient descent method, i.e.,  
\begin{align}
\theta_{k+1} = \theta_k - \alpha_k \nabla_{\theta} D^2(\pi_E, \pi_{\theta}, \mathcal{F})| _{\theta = \theta_k}
\end{align}
where $\theta_k$ denotes the parameters after updating $k$ iterations with initial policy $\theta_0$ and $\alpha_k$ denotes the learning rate. For the objective function
\begin{align}
D^2(\pi_E, \pi_{\theta}, \mathcal{F}) 
& =\| \mu_{\pi_{E}} -  \mu_{\pi_{\theta}}  \|^2_{\mathcal{F}} \nonumber \\
& =  \langle  \mu_{\pi_{E}}, \mu_{\pi_{E}} \rangle_{\mathcal{F}} + \langle  \mu_{\pi_{\theta}}, \mu_{\pi_{\theta}} \rangle_{\mathcal{F}} 
- 2  \langle  \mu_{\pi_{E}}, \mu_{\pi_{\theta}} \rangle_{\mathcal{F}},  \nonumber
\end{align}
only the last two terms contain the gradient information for $\pi_{\theta}$, and the gradient can be estimated by finite samples. 

Given $L$ trajectories of expert point processes and $M$ trajectories of events generated by $\pi_{\theta}$, we can have the finite sample estimate for $D^2(\pi_E, \pi_{\theta}, \mathcal{F})  $ by plugging in the mean embedding $\mu_E$ and $\pi_{\theta}$ as shown in Eq.~(\ref{eq:mean_embedding_est}), where 
\begin{align}
 &\langle  \mu_{\pi_{\theta}}, \mu_{\pi_{\theta}} \rangle_{\mathcal{F}} 
- 2  \langle  \mu_{\pi_{E}}, \mu_{\pi_{\theta}} \rangle_{\mathcal{F}}   
=  \frac{1}{M^2} \sum_{m=1}^{M}   \sum_{i = 1}^{N^{(m)}_T}\sum_{m'=1}^{M}   \sum_{j = 1}^{N^{(m')}_T} k(a^{(m)}_i,a^{(m')}_j) \nonumber \\
 &\quad\quad\quad\quad\quad\quad\quad - \frac{1}{ML} \sum_{m=1}^{M}   \sum_{i = 1}^{N^{(m)}_T}\sum_{l=1}^{L}   \sum_{j = 1}^{N^{(m')}_T} k(a^{(m)}_i,a^{(m')}_j).
\end{align}

The roll-out samples $a:=\{a_i\}$ are obtained by nonlinear transformations parametrized by $\theta$ and contain the derivative information for $\theta$. The gradient of the objective function is obtained as follows -- first take the derivative with respect to the actions, and then use the chain rule to take the derivative with the unknown policy parameters, i.e., 
\begin{align}
    \frac{\partial D^2(\pi_E, \pi_{\theta}, \mathcal{F})}{ \partial \theta}
    =  \frac{\partial D^2(\pi_E, \pi_{\theta}, \mathcal{F})}{ \partial a} \cdot \frac{\partial a}{ \partial \theta} .
\end{align}
where $a$ depends on the policy parameters through Eqs.~(\ref{eq:RNNmodel}) and~(\ref{eq:nonlinear}).

\end{document}